# Empirical Grounding Improves the Realism of LLM Agents Simulating Human Behavior During Disruptions

Chen Xia[1], Zexi Kuang[2], Yuqing Hu[3]*

[1] Department of Civil Engineering Technology, Environmental Management and Safety, Rochester Institute of Technology, Rochester, NY 14623, United States of America

[2] Khoury College of Computer Sciences, Northeastern University, Boston, MA, United States of America

[3] Digital and Interdependent Built Environment Laboratory, Department of Architectural Engineering, The Pennsylvania State University, University Park, PA, United States of America

**Corresponding author email:** yfh5204@psu.edu

**Abstract**

Large language model (LLM) agents offer a generative approach to simulating human behavior under conditions that may have few or no direct historical analogues, a common challenge in disaster and infrastructure-disruption planning. However, this generative capacity creates a validity problem: individually plausible agent reasoning may fail to reproduce empirical population behavior. We evaluate whether empirical grounding improves the statistical realism of LLM-agent simulations during disruptions. Specifically, we develop an empirically grounded LLM-agent framework that embeds demographic profiles from the American Community Survey, baseline routines from the American Time Use Survey, and urban spatial context into agent initialization, memory, decision prompts, and activity execution. An independent household survey conducted during the July 2024 Philadelphia heatwave is reserved as an external validation benchmark. Compared with an ungrounded LLM-agent baseline, the grounded model improved reconstruction of normal daily routines, increasing mean correlation with empirical activity profiles from 0.528 to 0.912 and reducing mean squared error from 0.066 to 0.008. Under heatwave conditions, the grounded model better reproduced survey-derived activity profiles, increasing mean correlation from 0.349 to 0.836 and reducing mean squared error from 0.098 to 0.012. The grounded model captured 46.4% of observed heatwave response amplitude, compared with 20.6% for the ungrounded baseline. These findings show that empirical grounding can make LLM agents more statistically credible simulators of population behavior while revealing remaining gaps in modeling human adaptation during disruptions.

**Significance Statement**

LLM agents are increasingly used to simulate human behavior in social, urban, and policy systems, but individual coherent reasoning does not ensure statistically realistic population behavior. This distinction matters when simulations inform planning, resource allocation, and climate adaptation decisions. The results show that grounding agents with empirical demographic and time-use data, together with urban spatial context, improves reproduction of observed activity distributions under normal and heatwave conditions. The grounded model also better captures the pattern of behavioral change from normal to heatwave conditions. Beyond the heatwave case, this paper suggests a path for using LLM agents to simulate future disruptions: empirical data anchor everyday behavior, while LLM reasoning generates adaptive responses when historical precedent is limited.

**Main Text**

Large language model (LLM)-based agents are emerging as a new approach for simulating human behavior in complex social and physical systems (1–3). Their distinctive promise is generativity: they can produce context-sensitive behavioral responses under conditions that may have few or no direct historical analogues. This capability is especially relevant for disaster and infrastructure-disruption planning, where future cities may experience compound hazards, cascading failures, or extreme events outside the range of past observations (22,26). Unlike conventional supervised learning models that depend on labeled observations from similar past events, LLM agents can combine agent profiles, memory, environmental information, and prompt-based reasoning to generate behavior under novel scenarios (1~3,23). However, this same generative capacity creates a fundamental validity challenge. Agents may produce actions that appear plausible at the individual narrative level while failing to reproduce the statistical structure of real population behavior (5,9,10). Recent studies have shown that LLM-generated social data may appear coherent but fail to reproduce empirical distributions, heterogeneous associations, and sequence-level patterns observed in real populations (4, 5). This concern is especially important for disruption simulation, where small distortions in individual behavior can aggregate into misleading estimates of exposure, shelter demand, mobility change, resource needs, and intervention effectiveness. For LLM agents to be useful in disaster and urban resilience research, their generated behavior must be evaluated not only by narrative coherence but also by statistical realism at the population level.

Empirical grounding offers a way to address this limitation, but its role in LLM-agent simulation remains underdeveloped. Human mobility, time-use, demographic, and spatial datasets provide evidence on how real populations organize behavior across time and space (6–8). These data can ground baseline routines and constrain feasible activity choices, but they usually describe realized behavior rather than adaptive decision-making under counterfactual or emerging disruptions. Conversely, LLM agents can generate adaptive responses under new conditions, but without empirical grounding, their outputs may reflect model priors or prompt assumptions rather than observed behavioral structure (9, 10). A central challenge, therefore, is how to combine empirical behavioral and spatial context with the generative flexibility of LLM agents.

Urban heatwaves provide a useful setting for evaluating this challenge because heat adaptation is often subtle, heterogeneous, and routine-constrained. Unlike hazards dominated by large-scale evacuation, heatwaves frequently lead to incremental behavioral adjustments, including changes in activity timing, time spent at home, and use of protective resources (12–15). Protective resources also matter, but their use depends on awareness, distance, and accessibility, which can limit the effectiveness of cooling centers and other public interventions (16). Evaluating heatwave behavior therefore requires more than testing whether agents react to heat; it requires assessing whether simulated populations reproduce observed shifts in activity distributions across time and social groups.

In this study, we evaluate whether empirical grounding can make the generative capacity of LLM agents more statistically credible for disruption simulation. We develop an empirically grounded LLM-agent framework that embeds behavioral context from American Community Survey demographic profiles and American Time Use Survey daily routines into agent initialization, memory, and decision prompts, while using the urban spatial environment as a constraint layer for activity execution (Fig. 1). Using a documented July 2024 Philadelphia heatwave as the testbed, we compare the proposed framework with an ungrounded LLM-agent baseline under normal and heatwave conditions. An independent household survey conducted during the heatwave is reserved as an external validation benchmark. The analysis evaluates whether grounded agents better reproduce baseline daily routines, heatwave-period activity profiles, and heatwave-induced behavioral shifts across demographic groups.

## Results

### Empirical grounding improves reconstruction of baseline activity routines

We first evaluated whether empirical grounding enables LLM agents to preserve realistic daily activity structure under normal conditions. This test assesses whether generative agents can reproduce baseline population routines when provided with empirical demographic and time-use context. For each demographic group, simulated activity probability profiles were compared with empirical baseline profiles derived from time-use data across the full time-by-activity matrix. Summary metrics were then averaged across demographic groups. The empirically grounded model reproduced the temporal structure of daily routines substantially better than the ungrounded LLM-agent baseline. Mean correlation with empirical activity profiles increased from 0.528 to 0.912, while mean squared error decreased from 0.066 to 0.008 (Table 1). Mean absolute error and Jensen-Shannon divergence showed the same pattern, indicating that empirical grounding improved both pointwise accuracy and distributional similarity. Across demographic groups, the grounded model consistently achieved higher temporal agreement with empirical activity profiles, whereas the ungrounded baseline showed larger distributional error. Full group-level validation metrics for normal-day simulations are reported in SI Appendix, Table S1.

A representative normal-day comparison is shown in Fig. 2 for employed adults without disability. This group was selected as an illustrative case because it has a clear workday structure, allowing direct comparison of routine timing across the empirical benchmark, the empirically grounded model, and the ungrounded LLM-agent baseline. The grounded model reproduced the expected daytime work pattern, the decline of biological needs after waking, and the evening increase in discretionary activity. In contrast, the ungrounded baseline produced a distorted activity profile in which biological needs dominated much of the day and work activity remained near zero. This failure mode illustrates the central validity problem for LLM-agent simulation: need-based reasoning can generate internally coherent individual actions while failing to reproduce the population-level temporal structure of routine behavior. Full normal-day activity comparisons for all demographic groups are provided in SI Appendix, Fig. S1-S9.

### Empirical grounding improves heatwave activity realism against an independent survey benchmark

We next evaluated whether empirical grounding improves LLM-agent simulation under heatwave disruption. Unlike the normal-condition test, which evaluates reconstruction of baseline routines, this stage uses an independent household survey conducted during the July 2024 Philadelphia heatwave as an external validation benchmark.Heatwave-day activity profiles from the empirically grounded model were compared with survey-derived heatwave activity profiles, while the ungrounded LLM-agent baseline was evaluated under the same environmental condition. Because the survey benchmark used broader activity and employment categories than the simulation model, heatwave validation was conducted using seven consolidated activity categories and five validation groups. The empirically grounded model remained substantially closer to observed heatwave activity profiles than the ungrounded LLM-agent baseline (Table 1). Mean correlation increased from 0.349 to 0.836, and mean squared error decreased from 0.098 to 0.012. Mean absolute error and Jensen-Shannon divergence also declined, indicating that empirical grounding improved both temporal alignment and distributional similarity under disruption. Across all five validation groups, the grounded model was closer to the survey-derived heatwave profile than the ungrounded baseline, whereas the ungrounded baseline failed to approximate observed heatwave activity structure in any group. Full group-level heatwave validation metrics are reported in SI Appendix, Table S2.

A representative heatwave-day comparison is shown in Fig. 3 for the employed group without disability. The empirically grounded model reproduced the broad temporal pattern of work or study activity, low daytime biological needs, and increased evening leisure activity. In contrast, the ungrounded LLM-agent baseline remained dominated by biological needs and showed limited sensitivity to the heatwave condition. These results indicate that empirical grounding improves not only baseline routine reconstruction, but also the ability of LLM agents to generate population-level activity distributions that align with observed behavior under heat stress. Full heatwave-day activity comparisons for all validation groups are provided in SI Appendix, Fig. S10-S14.

**Empirical grounding improves representation of heatwave-induced behavioral adaptation**
Finally, we evaluated whether agents reproduced observed behavioral adaptation from normal conditions to heatwave conditions. This evaluation is stricter than matching a single heatwave day profile because it asks whether the model captures the direction, structure, and magnitude of disruption-induced behavioral change. For each model, we calculated the heatwave-minus-normal change in activity probability and compared the resulting change magnitude with the empirical benchmark. The empirically grounded model captured a larger share of the observed response amplitude than the ungrounded LLM-agent baseline. Across demographic groups, the mean empirical behavioral change magnitude was 0.08156. The empirically grounded model produced a mean change magnitude of 0.03772, corresponding to 46.4% of the empirical response amplitude, whereas the ungrounded baseline produced a mean change magnitude of 0.01628, corresponding to 20.6% of the empirical response amplitude (Fig. 4A). Thus, empirical grounding more than doubled the share of observed heatwave adaptation captured by the simulation.

Category-level deltas further show how the models differed in their representation of heatwave adaptation (Fig. 4B). The empirical benchmark showed a decrease in biological needs and increases in health-related activity, household activity, work or study, and leisure. The empirically grounded model reproduced the dominant decrease in biological needs and the increase in leisure activity, but underestimated the full magnitude of observed adaptation. The largest remaining discrepancies occurred in work or study and travel or other activity, suggesting that additional behavioral mechanisms may be needed to represent institutionally mediated schedule adjustments and reduced nonessential travel under extreme heat. Full demographic-level change magnitudes and the category-level deltas are reported in SI Appendix, Tables S3 and S4

**Discussion**
This study shows that empirical grounding can substantially improve the statistical realism of LLM-agent simulations of human behavior. The broader contribution is not simply that empirical data improve simulation accuracy, but that empirical grounding helps make the generative capacity of LLM agents scientifically useful for disruption scenarios where direct historical observations may be limited or unavailable. Prior LLM-agent frameworks have demonstrated that agents can generate coherent individual behaviors, maintain memory, and reason within simulated environments (1–3). However, recent work has raised a central concern for social simulation: individually plausible outputs may still fail to reproduce empirical population-level distributions, heterogeneous associations, and sequence-level patterns (5). Our results support this concern and show a practical way to address it. Without empirical grounding, the ungrounded LLM-agent baseline produced activity profiles that were internally coherent but statistically unrealistic. By embedding demographic profiles and baseline time-use routines into agent memory and decision prompts, the empirically grounded model better reproduced both normal daily routines and heatwave-period activity profiles.

These findings clarify the relationship between generativity and validity in LLM-agent simulation. The value of LLM agents for disaster and infrastructure-disruption research lies in their ability to generate behavior under plausible future conditions, including events that a city has not previously experienced (23, 26). Yet generativity alone is not a validity criterion. An agent can produce behavior that sounds reasonable in natural language while still yielding aggregate activity distributions that are inconsistent with observed human behavior. Our results suggest that empirical behavioral grounding provides a necessary constraint: it anchors agents in realistic daily routines before allowing them to adapt under disruption. In this sense, empirical grounding does not replace LLM reasoning; rather, it gives LLM reasoning a statistically credible behavioral baseline from which adaptation can occur.

The findings suggest that empirical data should play a more active role in LLM-agent simulation than post hoc validation alone. Long-standing work on agent-based modeling has emphasized that simulation credibility depends on empirical validation, comparison with observed patterns, and careful assessment of whether micro-level rules produce credible macro-level outcomes (17, 18). This issue becomes more important for LLM-agent systems because natural-language reasoning can make individual actions appear plausible even when aggregate patterns are wrong (1,40). The framework developed here links empirical routine structure with LLM-based adaptive reasoning: empirical data establishes realistic baseline routines, while LLM agents can deviate from those routines under disruption. The normal-day results should therefore be interpreted as a routine-preservation and reconstruction test: they show whether the LLM-agent system can maintain empirically grounded behavioral structure during simulation. The heatwave results provide a stronger external validation test because the survey-derived heatwave activity profiles were not used to initialize, calibrate, or prompt the agents.

The heatwave results also show that empirical grounding improves, but does not fully resolve, the challenge of modeling behavioral adaptation. The grounded model captured a larger share of observed heatwave response amplitude than the ungrounded baseline and reproduced the dominant decrease in biological needs and increase in leisure activity. However, it underestimated the full magnitude of observed adaptation and showed remaining discrepancies in work or study and travel or other activity. This underestimation should not be interpreted only as a model limitation; it is also a substantive finding about what routine grounding can and cannot explain. Empirical time-use routines provide a strong baseline for everyday behavior, but heatwave adaptation also depends on institutional schedules, work flexibility, household resources, perceived risk, indoor thermal conditions, and constraints on nonessential travel (13, 24,25). Capturing these mechanisms likely requires additional empirically grounded adaptation modules rather than stronger prompting alone. This pattern is consistent with prior evidence that heat adaptation often occurs through subtle adjustments in activity timing, home-based behavior, and resource use rather than through a single protective action (12, 14). These results indicate that realistic heatwave simulation requires more than a generic heat response; it requires mechanisms that connect environmental stress with routine constraints, institutional schedules, and household-level coping options. This distinction is important for future applications that use LLM agents to evaluate disruption scenarios under realistic behavioral constraints.

The study also highlights the role of spatial context in empirically grounded LLM-agent simulation. In this paper, the urban environment is used primarily as a spatial constraint layer for activity execution, not as the main validation target. This distinction is important. Behavioral grounding constrains what agents are likely to do and when they are likely to do it, while the spatial environment constrains where those activities can feasibly occur. Future work can build on this structure by validating destination choice, travel distance, exposure locations, and protective-resource use against passive mobility data or facility-use records. Such extensions would allow LLM-agent simulations to move from statistically realistic activity profiles toward fully validated spatiotemporal behavior under disruption.

Several limitations remain. First, the heatwave benchmark is based on self-reported 24-hour activity data. Although the survey was conducted during the heatwave period, responses may contain recall error or subjective interpretation. Second, the heatwave survey supported broader employment categories than the simulation model. The simulation retained hybrid and in-person work distinctions, but validation was conducted at the broader employed-group level supported by the survey data. Future surveys should measure work mode directly to evaluate work-mode-specific adaptation. Third, the present validation focuses on group-level activity distributions rather than individual trajectories, causal reasoning, or destination-specific movement choices. Future work should incorporate sequence-level validation, passive mobility data, and point-of-interest-level comparisons. Finally, heat adaptation is shaped by indoor thermal conditions, household energy constraints, and building characteristics that are not fully represented here. Integrating building thermal performance and indoor exposure estimates would allow future LLM-agent simulations to link behavioral adaptation more directly to heat-health risk.

Overall, this study establishes empirical grounding and external validation as necessary conditions for using LLM agents as credible simulators of population behavior under disruption. LLM agents are valuable because they can generate behavior under novel and uncertain conditions, but their outputs should not be accepted on narrative plausibility alone. By combining empirical behavioral routines, demographic profiles, spatial constraints, and independent survey validation, this study provides a path toward LLM-agent simulations that are not only generative, but also statistically testable and behaviorally realistic.

## Materials and Methods

### Study setting and experimental design

We used Philadelphia, Pennsylvania, as an empirical testbed to evaluate whether empirical grounding improves the statistical realism of LLM-agent simulation under disruption. The simulation compared normal summer weekday conditions with a documented July 2024 heatwave period (27). Normal conditions were used to evaluate whether agents could preserve and reconstruct empirically grounded baseline routines, while heatwave conditions were used to evaluate whether agents produced activity changes consistent with observed behavioral adaptation. The model generated individual activity states, activity transition times, and spatial locations, which were aggregated into group-level activity probability profiles for validation.

The experimental design compared two agent systems. The first was an ungrounded LLM-agent baseline based on the original AgentSociety framework, which represents behavior primarily through generic need-based reasoning (3). The second was our empirically grounded framework, which embeds demographic profiles and baseline time-use routines into agent initialization, memory, and decision prompts. This comparison isolates the contribution of empirical grounding while keeping the broader simulation environment consistent across models.

### Empirical grounding of baseline routines

Baseline routines were constructed from the American Time Use Survey and demographic information from the American Community Survey (19,20). The time-use data provided 24-hour activity diaries, while the demographic data were used to assign agents to population groups in the Philadelphia study area. Agents were stratified by employment status, work mode, and household or physical constraint indicators (28~31). Full population segmentation and sample mapping are reported in SI Appendix Method S1. Activities were represented using major daily activity categories. For normal-condition validation, activity profiles were organized into eight categories: health, biological needs, household management, personal obligations, work, education, personal preference, and other movement-related activities. The full recoding from original time-use activity codes is reported in SI Appendix Method S1. For each demographic group, time-varying activity transition patterns were estimated from the time-use data. A Markov-chain procedure based on our previous research generated stochastic daily activity sequences at hourly resolution (21). The resulting routines provided each agent with a baseline behavioral schedule and a time-specific activity probability profile. These routines served as behavioral context for the LLM agent rather than fixed rules that fully determined behavior.

These empirically derived routines were used as behavioral context rather than deterministic rules. During simulation, agents could deviate from their baseline routines when prompted by environmental conditions, recent activity history, or need states. This design allowed empirical data to anchor normal behavior while preserving the LLM agent's capacity to generate adaptive responses under heatwave conditions.

**LLM agent simulation framework**

The simulation framework was built on AgentSociety, a large-scale LLM agent platform that represents agents with profile memory, state memory, and stream memory (3). In the original framework, activity decisions are generated primarily through general need-based mechanisms. We modified this framework by inserting empirically derived demographic attributes and baseline activity routines into agent memory and decision prompts.

Each agent was initialized with a profile containing demographic and role-related information. The agent state also included a typical daily activity profile derived from the corresponding time-use group. During simulation, the LLM agent received prompts that combined the agent profile, recent activity history, and current environmental condition. The prompt instructed the agent to follow its empirical routine unless the current context created a reason to adapt. In this way, heatwave behavior was modeled as a context-sensitive deviation from a realistic baseline routine rather than as a response generated from generic heat-related assumptions alone.

The decision process focused on activity selection and execution. At each simulation step, the agent evaluated its current need state and environmental context. The planning module then selected a high probability activity category and translated it into executable actions. The grounded model differed from the baseline by explicitly including the empirical activity probability for the current time period in the decision context. Full agent architecture, internal thresholds, and implementation parameters are reported in SI Appendix, Method S2.

**Spatial environment and activity execution**

The simulation embedded agents in a spatial representation of central Philadelphia. The spatial environment was constructed using OpenStreetMap road and building data (32), SafeGraph point of interest data (33), and the Moss/Mosstool simulation toolkit (34). Road networks, building footprints, and activity-area footprints are extracted for central Philadelphia within the study bounding box (39.90°N–40.00°N, −75.24°W to −75.09°W) and transformed using a transverse Mercator projection to maintain metric consistency for movement simulation. The processed geographic layers are serialized into a Protocol Buffer–based map representation compatible with the AgentSociety simulation engine. The map provided feasible locations for home, work, and other activity destinations. The spatial environment was used as a constraint layer rather than as the primary object of inference. The main outcomes analyzed in this paper were activity probability profiles and activity changes under heatwave conditions.

**Independent heatwave survey benchmark**

Heatwave adaptation was validated using an independent household survey conducted in Philadelphia during the July 2024 heatwave period. The survey was administered during consecutive National Weather Service heat warnings and retained 603 valid responses after quality screening. Respondents reported 24-hour activity and time-use information during the heatwave period. These responses were used only as an empirical validation benchmark, not as an input for initializing, calibrating, or prompting the agents. Because the survey benchmark used broader categories than the simulation routine model, heatwave validation used seven consolidated activity categories. Work and education were combined into a single work or study category because both represent structured activities shaped by external schedules. Employment-related groups were also merged where needed to avoid over-fragmenting the empirical benchmark. The simulation retained the finer work-mode distinctions, but validation was conducted at the coarser grouping supported by the survey data.

The human-subjects survey was reviewed and approved by the Pennsylvania State University Institutional Review Board under STUDY00025094. Eligible participants were adults aged 18 years or older residing in the Philadelphia DMA area. Participants provided electronic informed consent before completing the online survey; the consent form was presented as the first page/question of the survey, and participants had to indicate consent before proceeding.

**Validation metrics**

Validation was performed in two stages. First, normal-condition outputs were compared with empirical baseline activity profiles derived from time-use data. Second, heatwave-condition outputs were compared with survey-derived heatwave activity profiles. These two stages evaluated whether the model reproduced routine population behavior and disruption-period activity patterns. For each demographic group, simulated and empirical activity probability matrices were compared across time and activity categories. Pearson correlation measured temporal alignment, mean squared error and mean absolute error measured numerical deviation, and Jensen-Shannon divergence measured distributional distance (35~37). Using multiple metrics allowed us to distinguish temporal-pattern agreement, pointwise error, and distributional similarity. To evaluate heatwave adaptation, we compared changes from normal to heatwave conditions. Change magnitude was calculated as the mean absolute heatwave-minus-normal activity shift across demographic groups and activity categories. Category-level deltas were also computed to identify which activities increased or decreased under heatwave conditions (38,39). This adaptation analysis provided a stricter test than heatwave-profile matching alone because it evaluated whether the model captured the direction and magnitude of behavioral change induced by heatwave conditions.

**Acknowledgments**

This research was supported by a seed grant from the Penn State Center for Socially Responsible Artificial Intelligence (CSRAI). The findings and conclusions are those of the authors and do not necessarily represent the views of CSRAI or Penn State.

**Figures and Tables**

**Figure 1.**

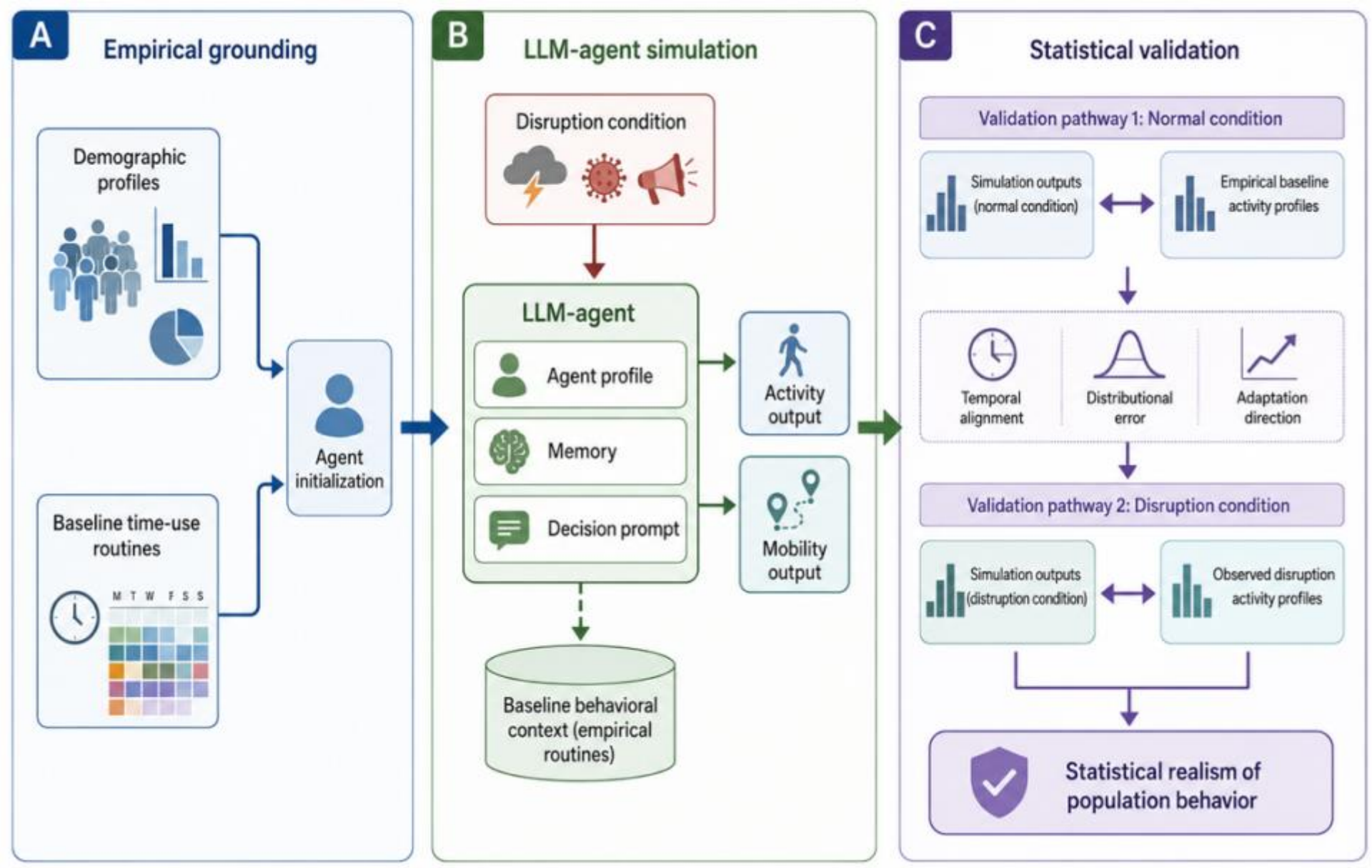

Figure 1. Framework for empirically grounded LLM-agent simulation and validation

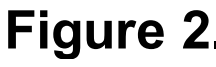

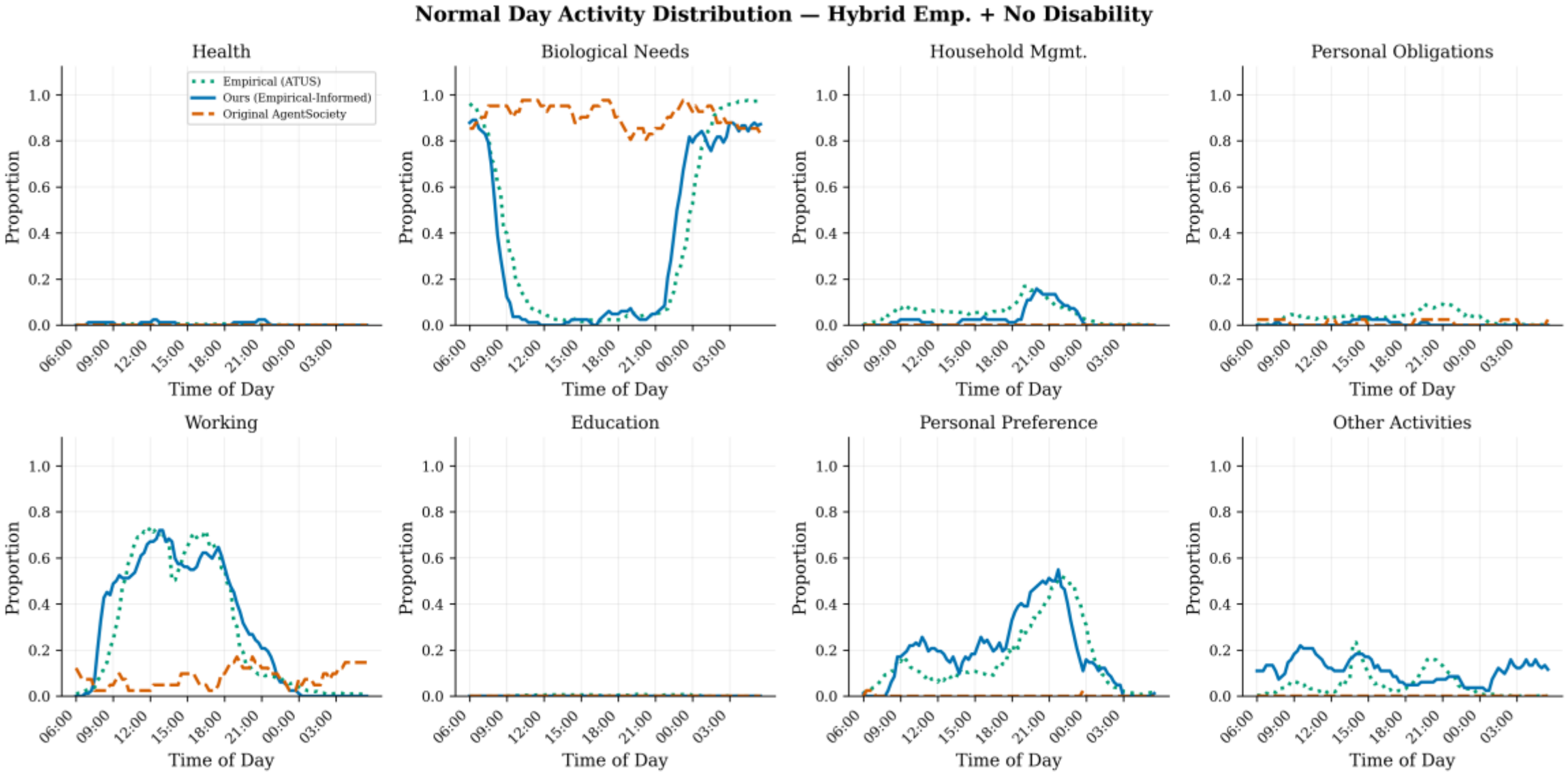

Figure 2. Empirical grounding improves reconstruction of normal-day activity routines.

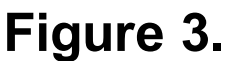

**Figure 3.**

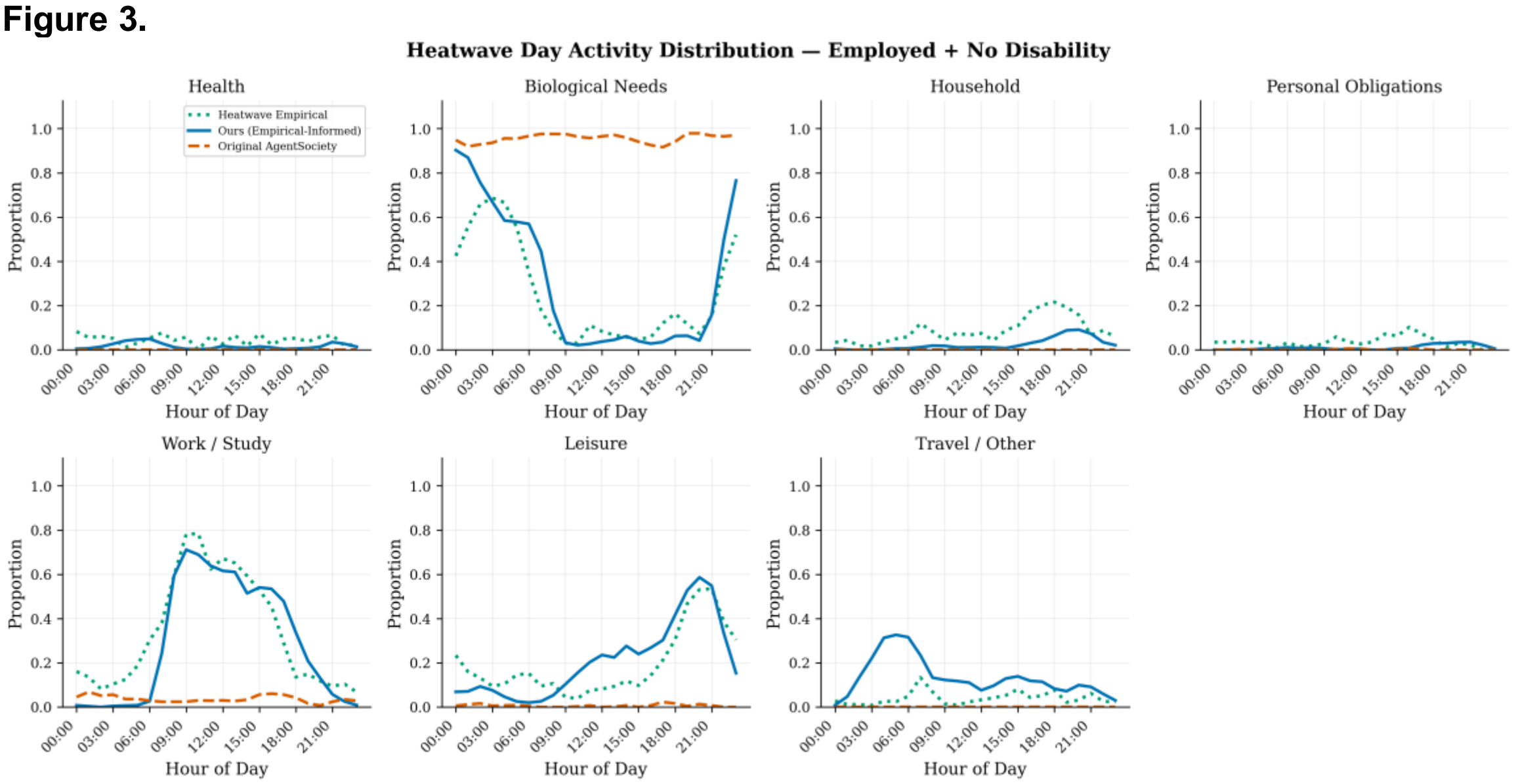

Figure 3. Empirical grounding improves heatwave activity realism.

**Figure 4.**

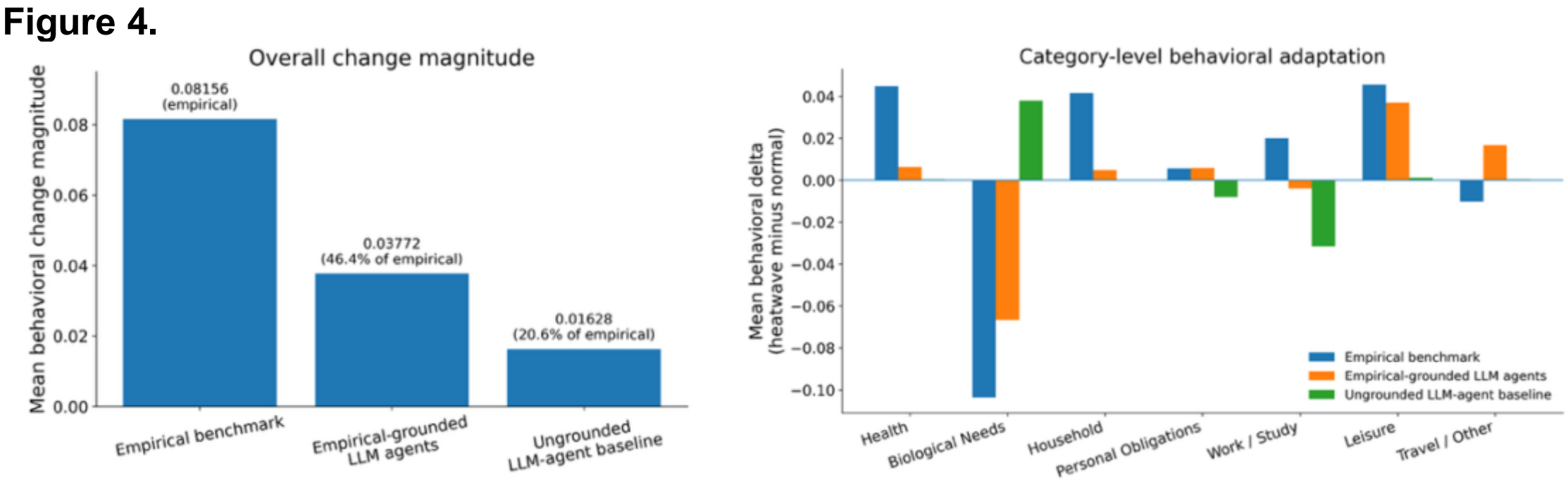

(A) Overall change magnitude (B) Category-level behavioral adaptation

Figure 4. Heatwave-induced behavioral adaptation.

**Table 1.**

Summary validation metrics for empirically grounded and ungrounded LLM-agent simulations

| Evaluation target | Model | MSE | MAE | Corr | JSD |
|---|---|---|---|---|---|
| Normal activity routines | Empirically grounded | 0.008344 | 0.055411 | 0.911789 | 0.008455 |
| Normal activity routines | Original LLM-agent | 0.065990 | 0.131900 | 0.528289 | 0.033079 |
| Heatwave activity profiles | Empirically grounded | 0.012335 | 0.080540 | 0.836360 | 0.013261 |
| Heatwave activity profiles | Original LLM-agent | 0.097867 | 0.189500 | 0.348680 | 0.046506 |

**Supporting Information for**

# Empirical Grounding Improves the Realism of LLM Agents Simulating Human Behavior During Disruptions

Chen Xia, Zexi Kuang, Yuqing Hu*

Email: yfh5204@psu.edu

**This PDF file includes:**

## Supporting Information Text

### Method S1 Constructing Demographically Informed Daily Routines

To capture realistic, diverse mobility patterns, we construct daily routines that are demographically informed using the American Time Use Survey (ATUS) and American Community Survey (ACS). These datasets provide complementary information: ATUS offers 24-hour diaries of individual activities with timestamped detail, while ACS provides the demographic composition of geographic areas, including age, occupation, household structure, and income levels. First, we select population segments most relevant to the heatwave in Philadelphia. We represent heterogeneity with nine agent groups defined by three behavior-relevant attributes: (i) employment & work mode, (ii) disability, and (iii) presence of children <18 in the household. These attributes directly shape mandatory activities (e.g., work or caregiving) and time–location constraints. For Employment & work mode, paid work is the dominant "mandatory" activity shaping start times, location (home vs. workplace), and commuting. Hybrid/telework measurably shifts time-at-home and commute incidence (1,2). For personal physical condition, disability is associated with different activity repertoires, shorter/less frequent trips, and higher home-stay time (3). For household obligations, care obligations introduce school/childcare drop-off windows, elevate shopping/errand frequency, and compress discretionary time (4). When attributes overlap (e.g., children and disability), we prioritize the condition most likely to alter time constraints and encode the other as a modifier in duration and travel parameters. Table S5 shows the nine groups of our agents with detailed demographic combination and key routine constraints during weekdays.

We construct daily routines for each agent group by processing microdata from the 2023 American Time Use Survey. We prioritized the 2023 dataset as it provides the most recent complete annual record proximate to our July 2024 case study. Crucially, this period captures stabilized 'new normal' behavioral patterns, particularly the persistence of hybrid work and altered commuting rhythms, that have emerged following the volatility of the COVID-19 pandemic. This ensures our agents reflect current urban dynamics rather than outdated pre-pandemic routines or temporary crisis-response behaviors seen in 2020–2022. Respondents are mapped to the predefined demographic–occupational agent types. In this context, "activity" denotes a sequence of behaviors undertaken over time. For major activity on daily routing, we adopt the main eight categories based on existing research that captures most of activities in ATUS (5) : (1) essential health activities, (2) biological needs (eating, sleeping), (3) working, (4) education, (5) household management, (6) personal obligations (shopping, banking, caregiving, etc.), (7) personal preference (leisure/social/recreation), and (8) travel/other movement. We start with ATUS's 18 major activity codes and categorize them according to functional similarity to the proposed activity categories. For example, "socializing," "relaxing," and "sports" are grouped under personal preference to reflect their discretionary nature.

Then, we simulate daily activity sequences using a Markov chain model, a widely adopted approach for modeling categorical time-series data in activity-based travel and behavior research (6,7). This method is particularly suited to representing stochastic human behavior because it captures both the sequential dependence between activities and the inherent variability across individuals. Formally, a first-order Markov chain assumes that the probability of transitioning to a given activity state at time $t$ depends only on the state occupied at $t-1$. To incorporate diurnal variation, we estimate a separate transition probability matrix for each discrete time interval (15 min), stratified by agent group and day type (this research focuses on weekday patterns). To better reproduce empirical behavior patterns, we extend the basic Markov framework with semi-Markov dwell-time constraints, sampling minimum durations for each activity from the empirical ATUS distributions. This adjustment prevents unrealistically short activity fragments, such as work periods of only a few minutes, while preserving stochastic variability. Scaling the simulation to large agent populations ensures convergence of the group-level probability profiles while maintaining diversity in individual schedules. For this study, the result then aggregated into hourly resolution daily routines whose statistical properties align closely with observed distributions, providing a realistic and flexible behavioral baseline for subsequent hazard-response modeling.

**Method S2 Agent Architecture and Empirical Data Integration**
We build our framework based on AgentSociety framework, an architecture designed for large-scale urban simulation with LLM-powered reasoning (8). First, we used the memory-from-file initialization method available in AgentSociety. Instead of random attributes, each agent's profile and state are from individual-level records derived from the American Community Survey (ACS). We construct 369 agents distributed across 41 census tracts for Philadelphia, each with home and work locations (mapped to Area of Interest (AOI) IDs in the urban environment), along with demographic attributes. These attributes are directly incorporated into agents' Profile and State memory during initialization. Specifically, for an agent from a given population segmentation (G1-G9), during initialization, we assign them a typical daily schedule that specifies the most probable activity for each time slot and the full probability distribution based on empirical data. This schedule is stored in the agent's State memory as typical_daily_activities, providing an empirically grounded and searchable baseline routine against which disruptions can be measured. Fields such as age, occupation, and household relationships are marked for embedding, enabling semantic retrieval during decision-making. When an agent receives a prompt to decide whether to evacuate, seek cooling, or continue working, the LLM queries memory for relevant profile attributes and past activities, which are then concatenated into the reasoning context. This retrieval-augmented prompting ensures that decisions reflect the agent's empirical characteristics rather than generic assumptions.

For agent’s actions, AgentSociety employs a block dispatcher that selects task-specific reasoning modules (blocks) based on agent intentions. Each block constructs a prompt that combines the agent's memory, current context, and task objectives, then queries the LLM for a response. For this project we focus on two blocks, which are NeedsBlock and PlanBlock. Together, they generate the agent's activity decisions. We modify these two blocks to explicitly incorporate empirical attributes and time-use baselines,

At each simulation step, the NeedsBlock first assesses four satisfaction dimensions, including hunger, energy, safety, and social, which decay over time. When a satisfaction value drops below its threshold (0.2 for hunger, energy, and safety; 0.3 for social), the corresponding need is prioritized. All four dimensions are held constant while the agent is executing a sleep plan, and satisfaction is restored upon completing the relevant activity. External interventions, such as heatwave warnings, can trigger need reassessment through a reflection mechanism where the LLM evaluates whether the agent should abandon its current plan.

Given the prioritized need, the PlanBlock then selects a high-level guidance from nine activity options aligned with our ATUS categories (sleeping, health, eating, household management, personal obligations, working, education, personal preference, other activities). The selection prompt incorporates the agent's demographic profile (age, occupation, income, work mode), current environmental conditions (weather, temperature, announcements), the ATUS-derived activity probability distribution for the current time slot, and the agent's recent activity history over the past 4 hours to penalize repetition. Perceived Behavioral Control is replaced by Consistency with Typical Behavior (alignment with the empirical activity distribution, weighted as the most important factor) to anchor decisions in the agent's demographic baseline; and a Repetition Penalty is added to suppress short-horizon cycling. The prompt instructs the agent to follow its daily routines unless urgent needs or special circumstances arise. This design ensures that agents default to empirically realistic behavior under normal conditions while retaining the capacity for adaptive deviation during emergencies.

The selected guidance is then expanded into a sequence of executable steps, each typed as one of four action categories: mobility, social, economy, or other. A second LLM call evaluates the plan on goal alignment, completeness, feasibility, and time appropriateness. If the overall score falls below 0.5, the plan is regenerated for up to two additional attempts before the best available plan is accepted. This loop enforces consistency. For example, a work plan for an in-person employee at home must include a commute step, and a sleep plan must include an actual sleep action rather than only preparation. Agents are classified by work mode based on their demographic group: in-person employees must commute to their workplace, hybrid employees can work remotely, and unemployed agents have no work obligations. This distinction is injected into both the guidance selection and plan generation prompts, directly affecting spatial mobility patterns.

## Figures

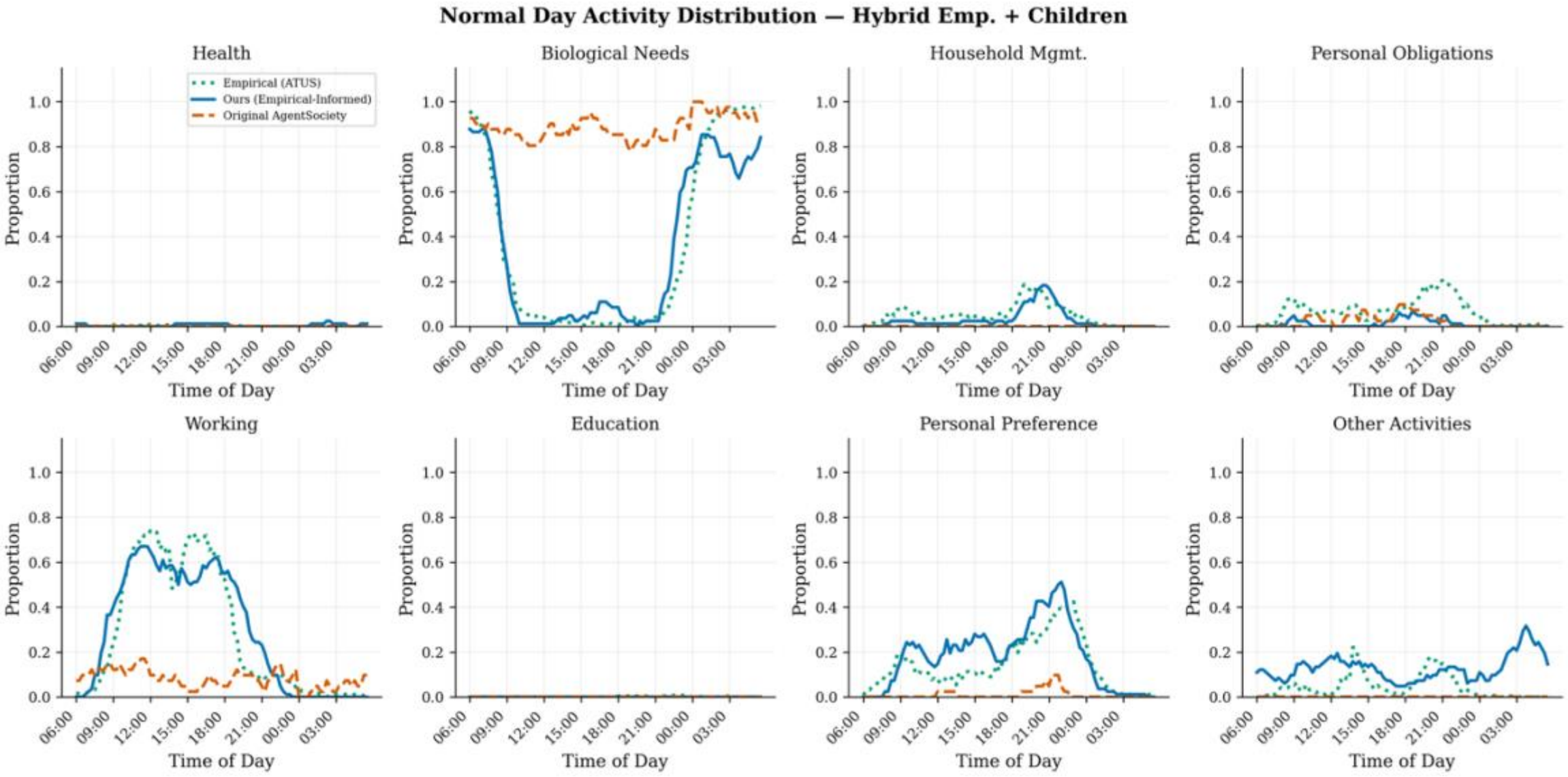

**Fig. S1.** Normal-day activity comparison for hybrid employed agents with children

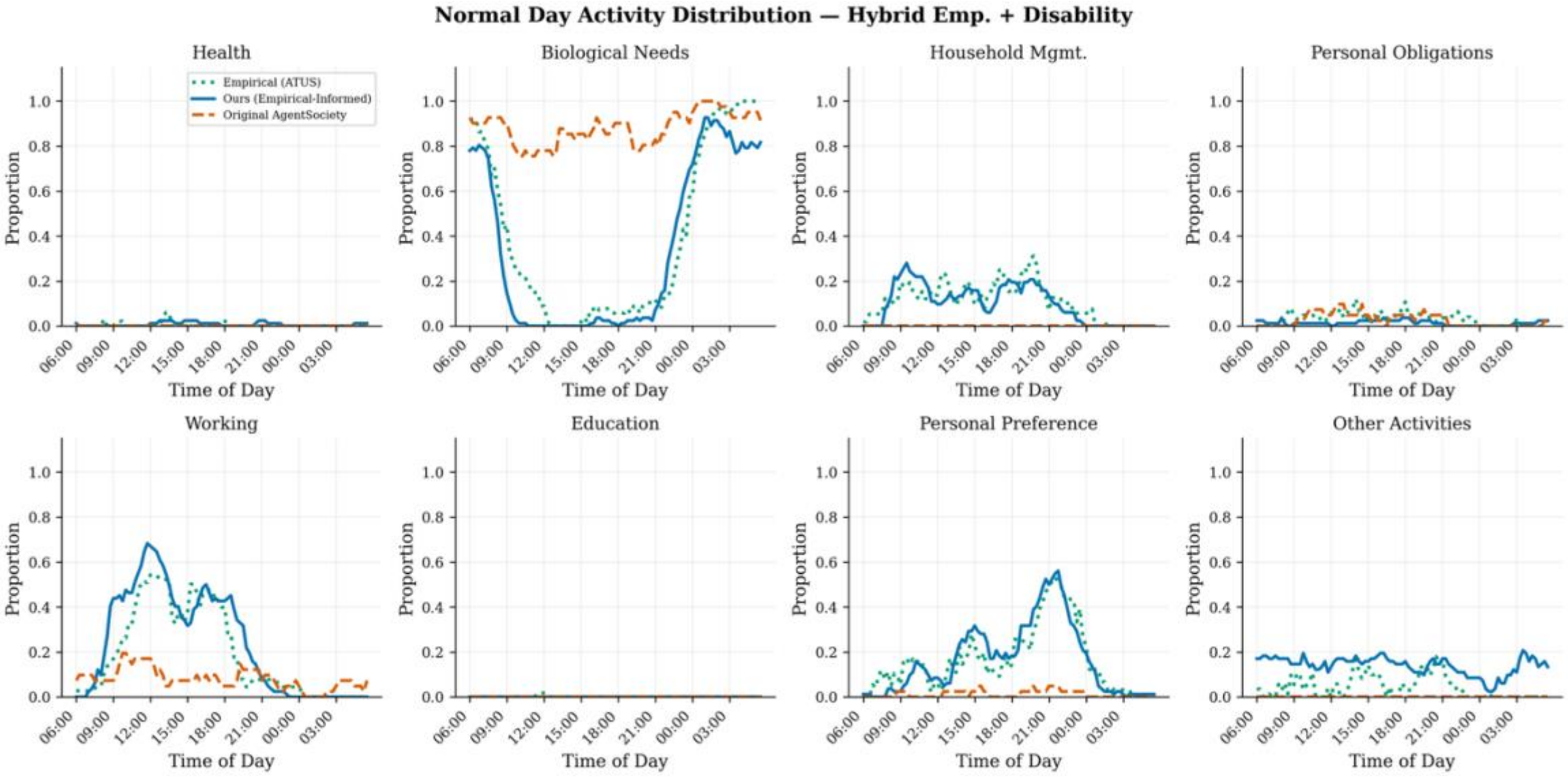

**Fig. S2.** Normal-day activity comparison for hybrid employed agents with disability.

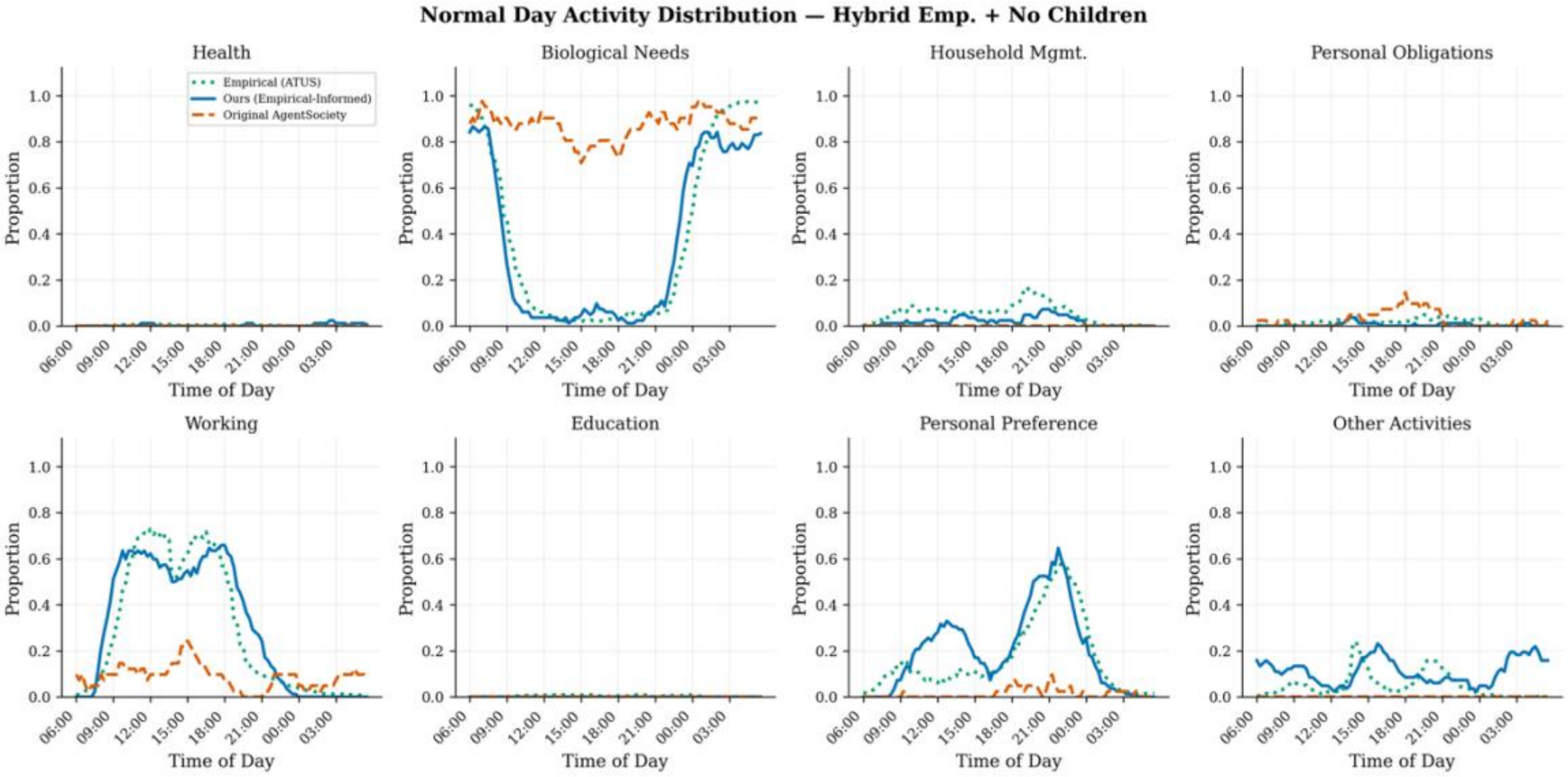

**Fig. S3.** Normal-day activity comparison for hybrid employed agents without children.

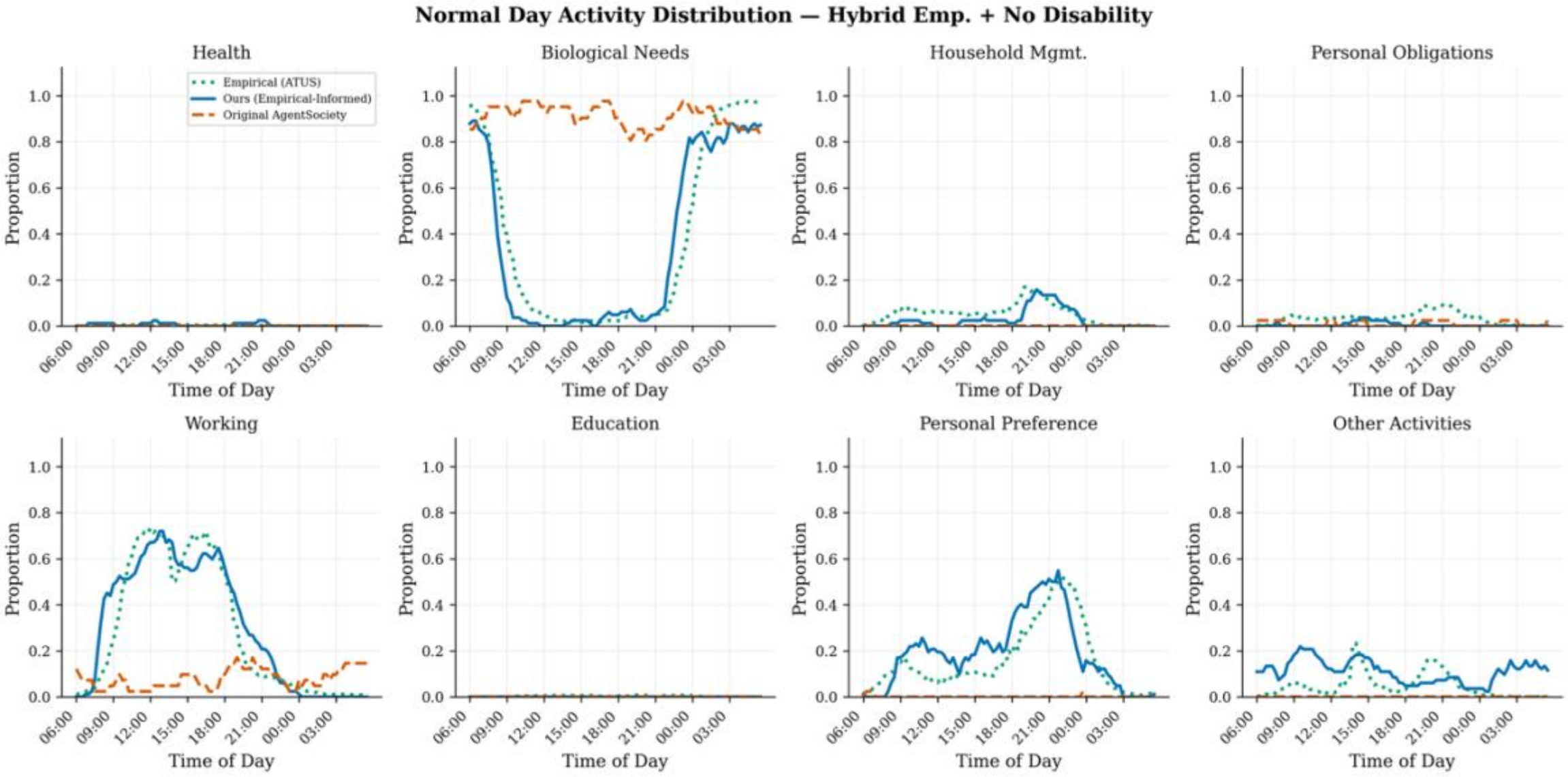

**Fig. S4.** Normal-day activity comparison for hybrid employed agents without disability.

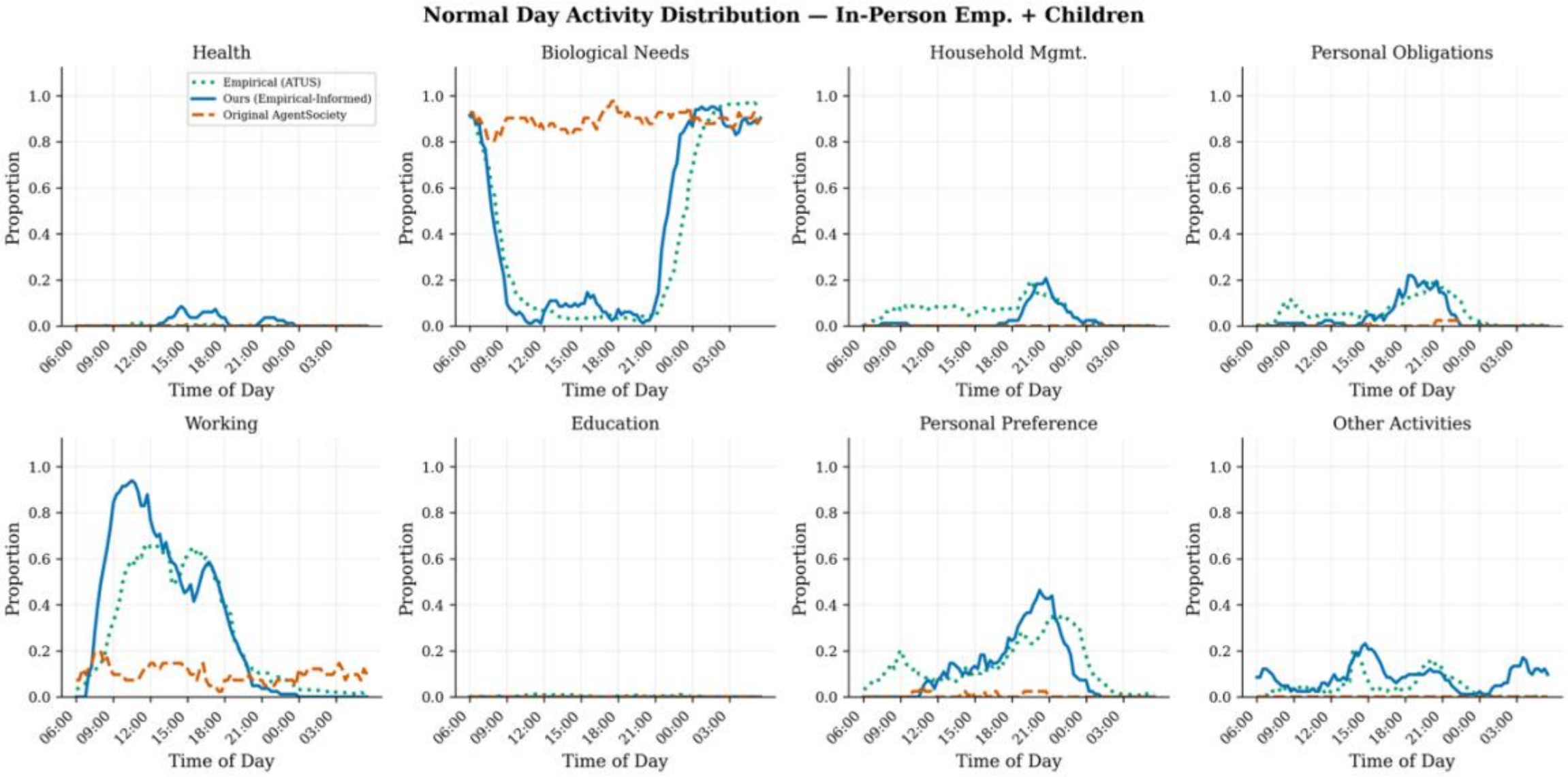

**Fig. S5.** Normal-day activity comparison for in-person employed agents with children.

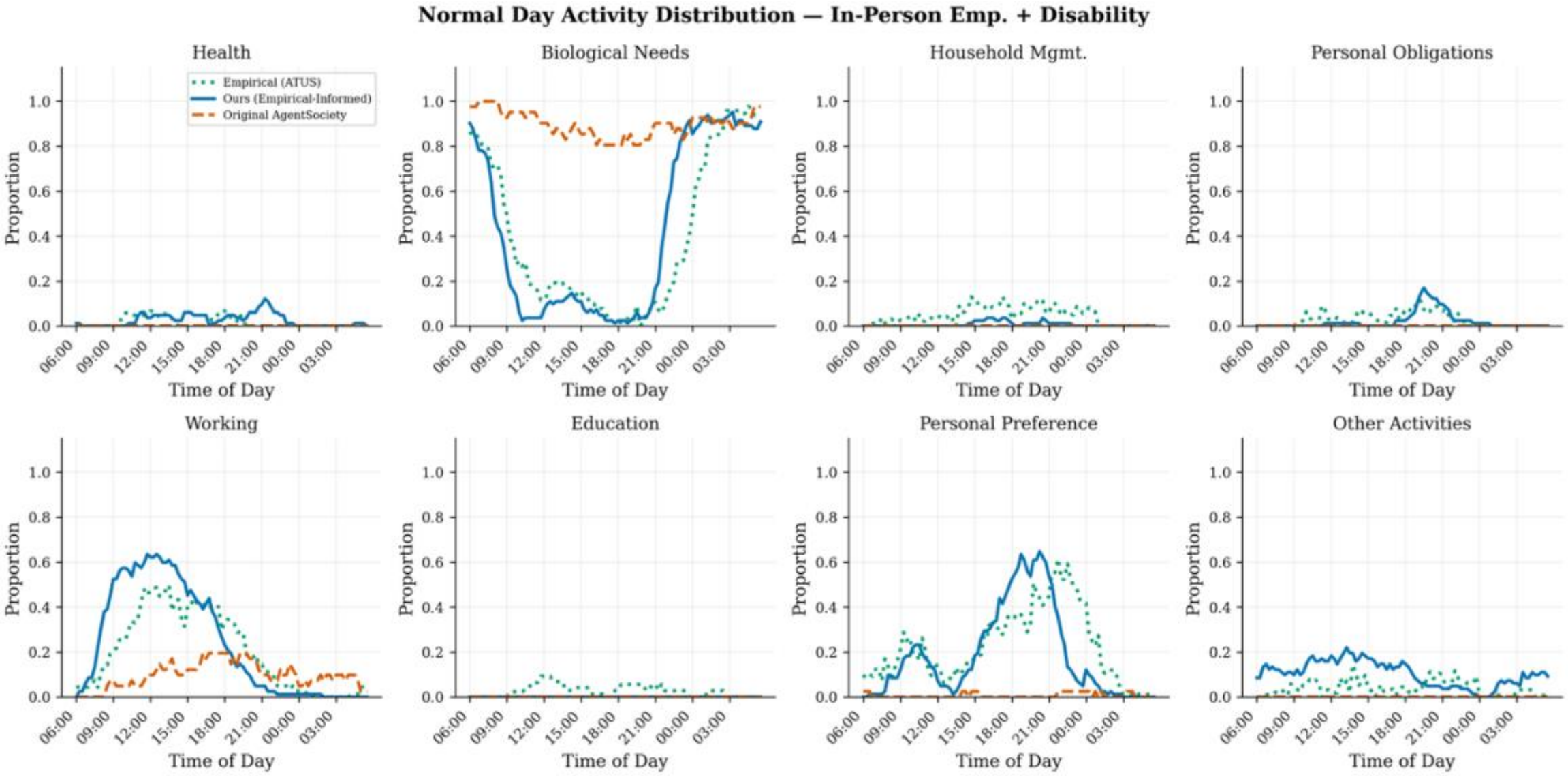

**Fig. S6.** Normal-day activity comparison for in-person employed agents with disability.

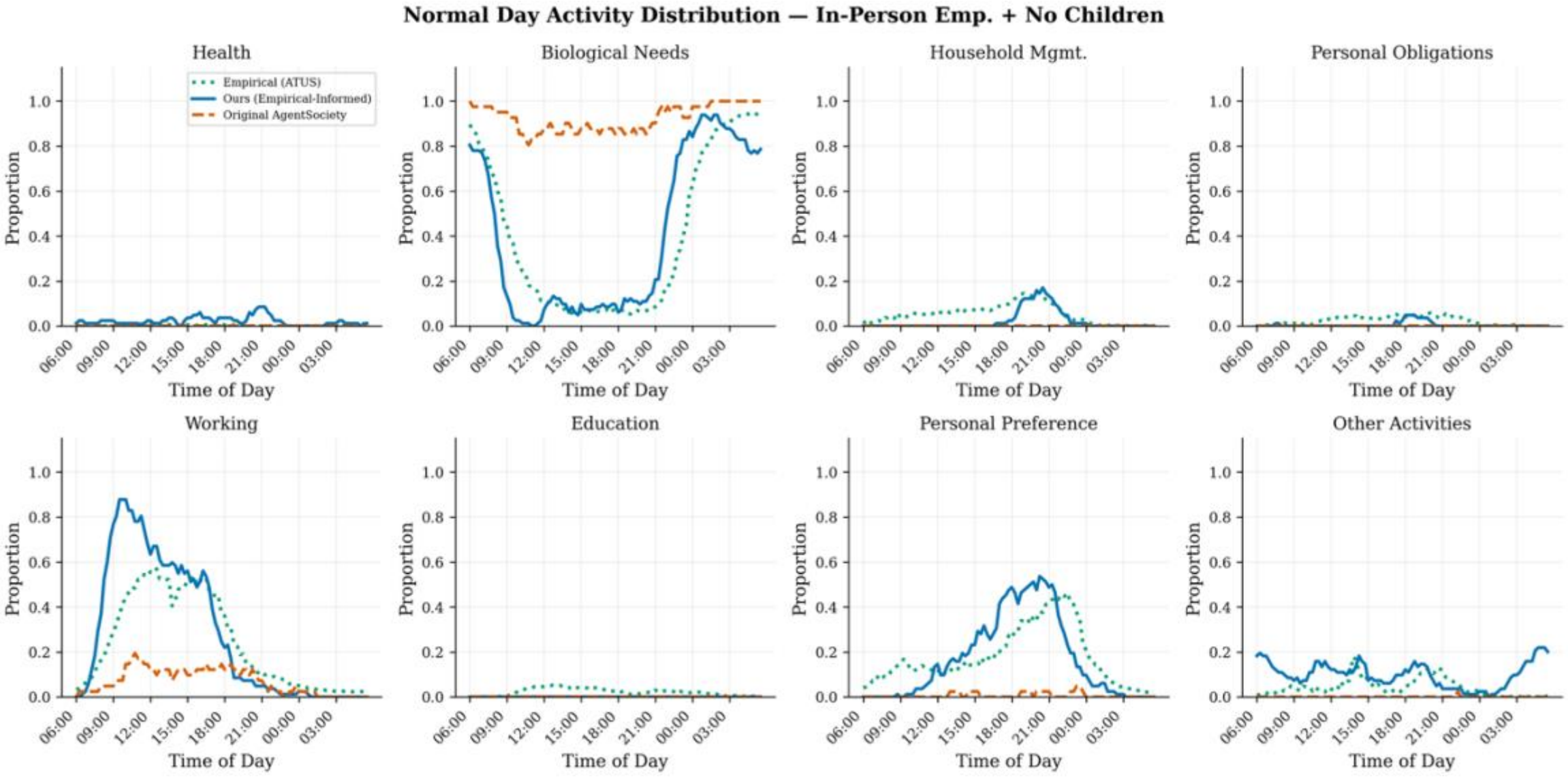

**Fig. S7.** Normal-day activity comparison for in-person employed agents without children.

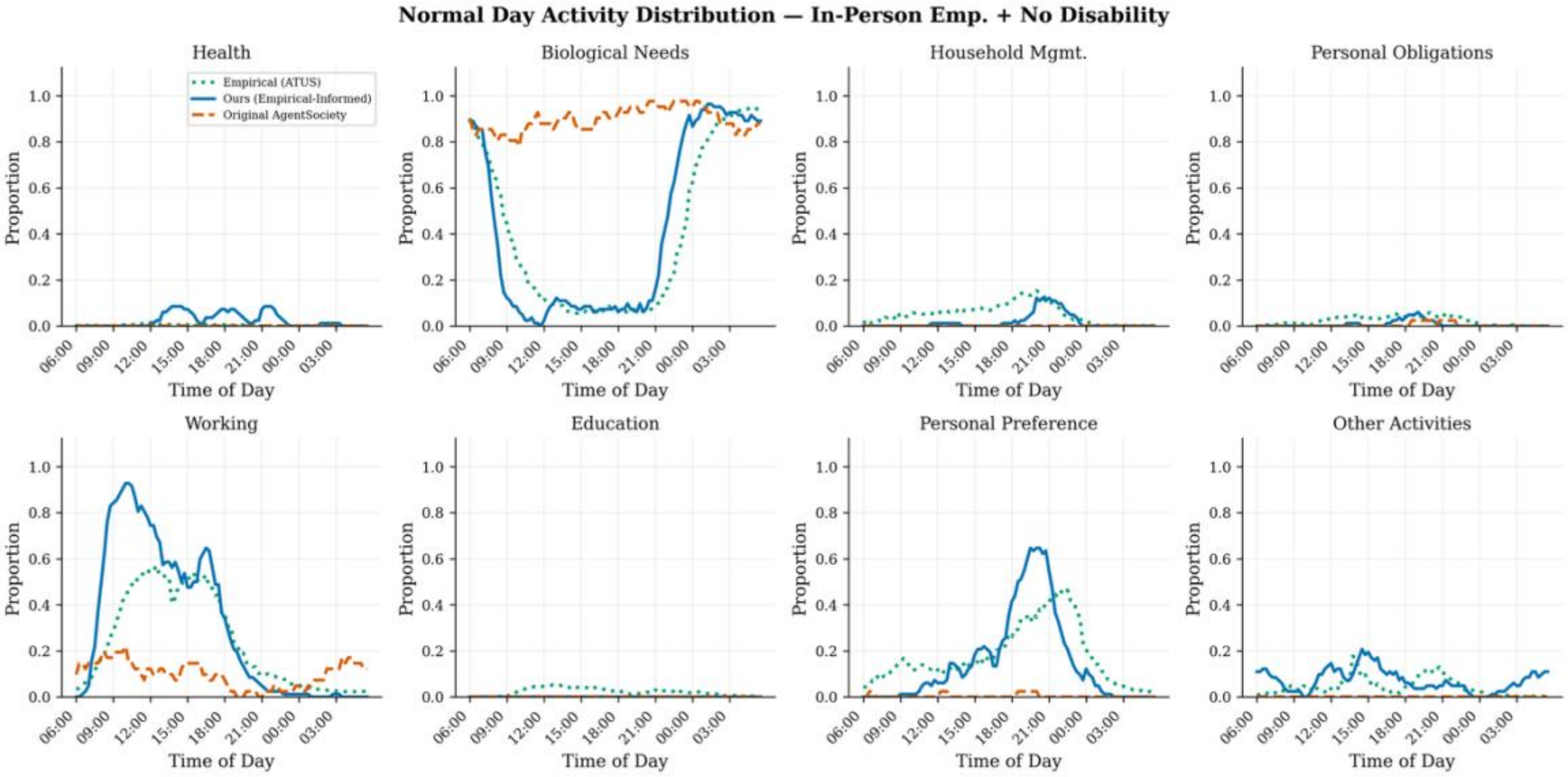

**Fig. S8.** Normal-day activity comparison for in-person employed agents without disability.

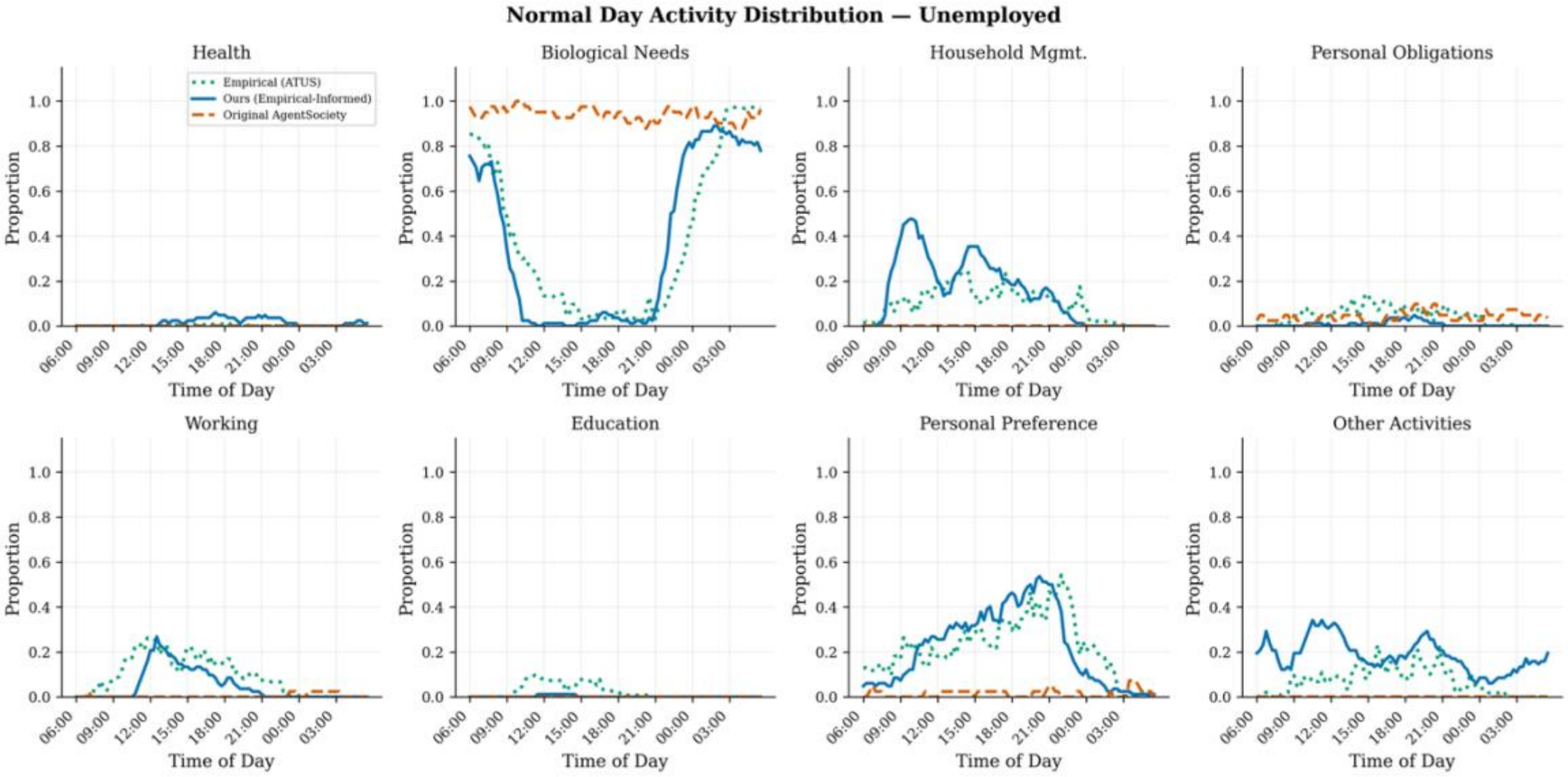

**Fig. S9.** Normal-day activity comparison for unemployed agents.

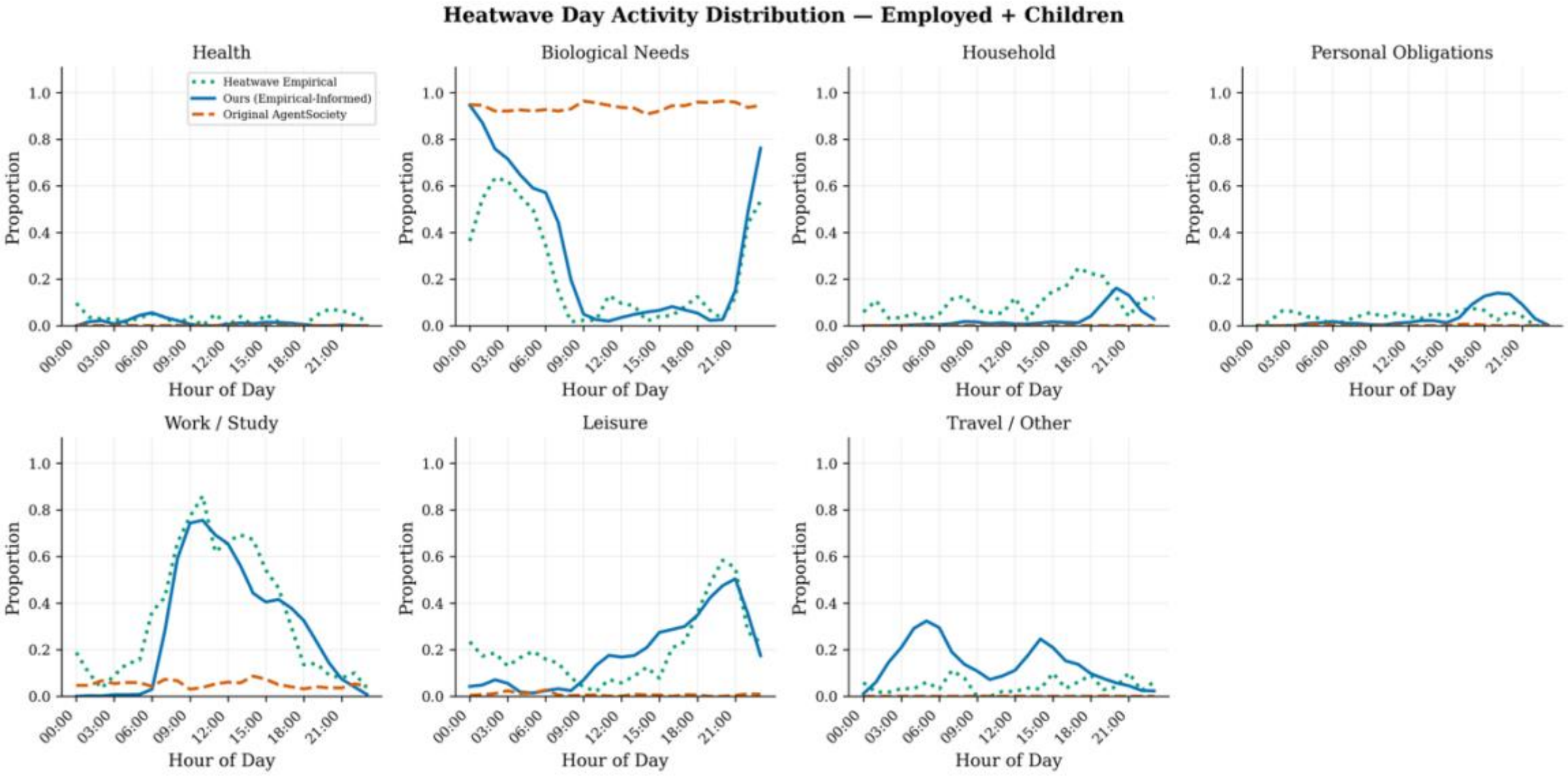

**Fig. S10.** Heatwave-day activity comparison for employed agents with children.

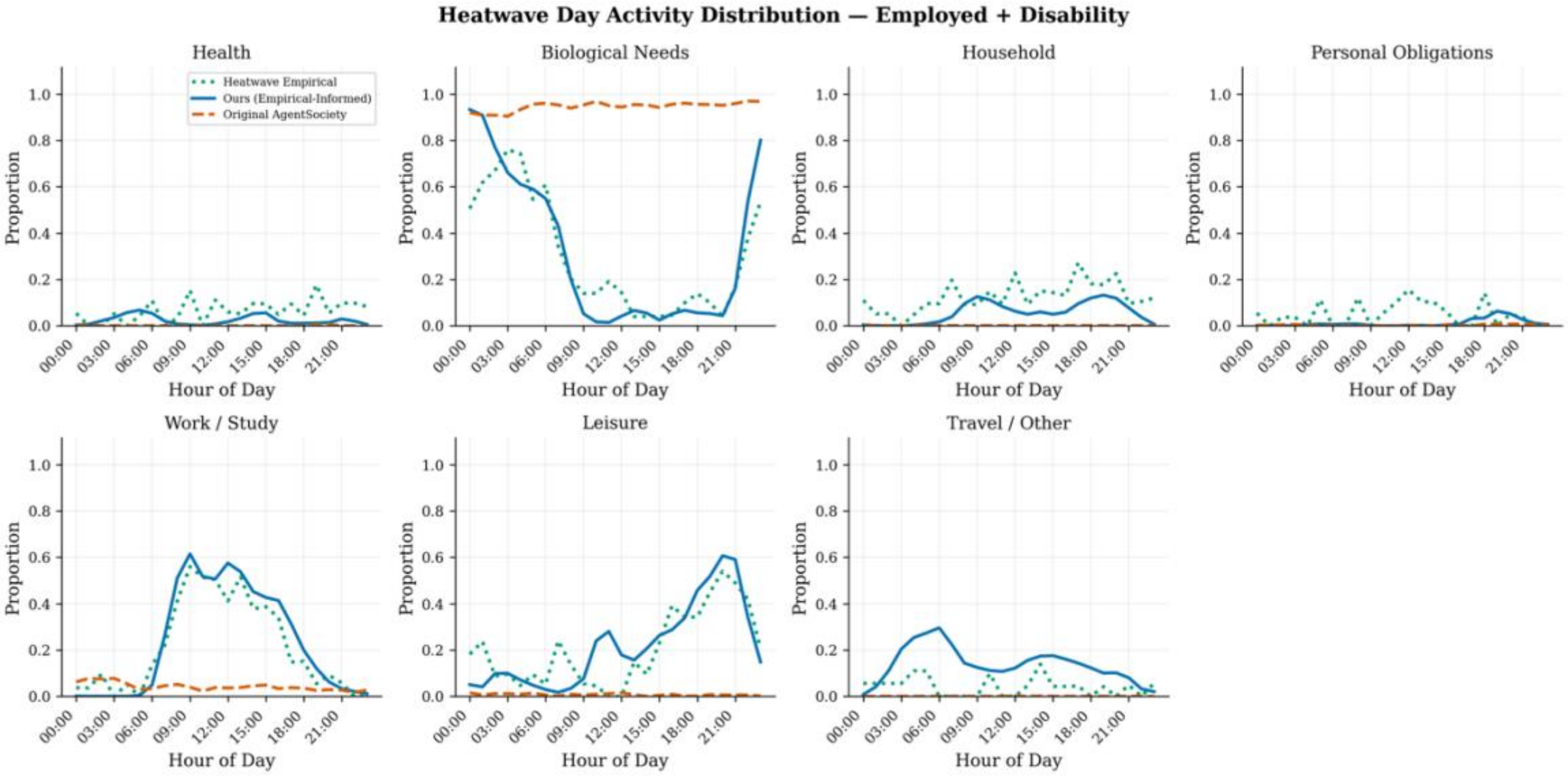

**Fig. S11.** Heatwave-day activity comparison for employed agents with disability.

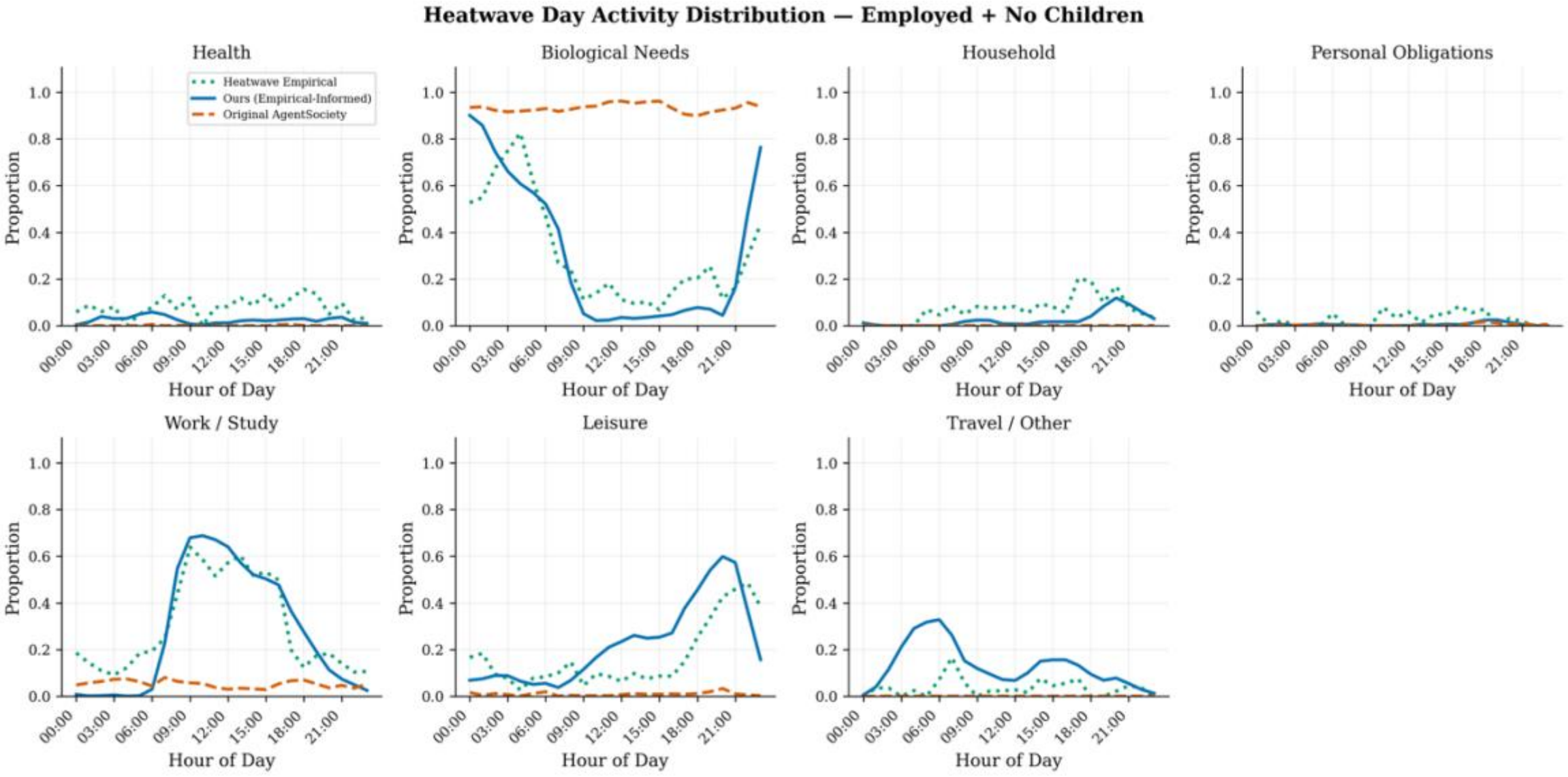

**Fig. S12.** Heatwave-day activity comparison for employed agents without children.

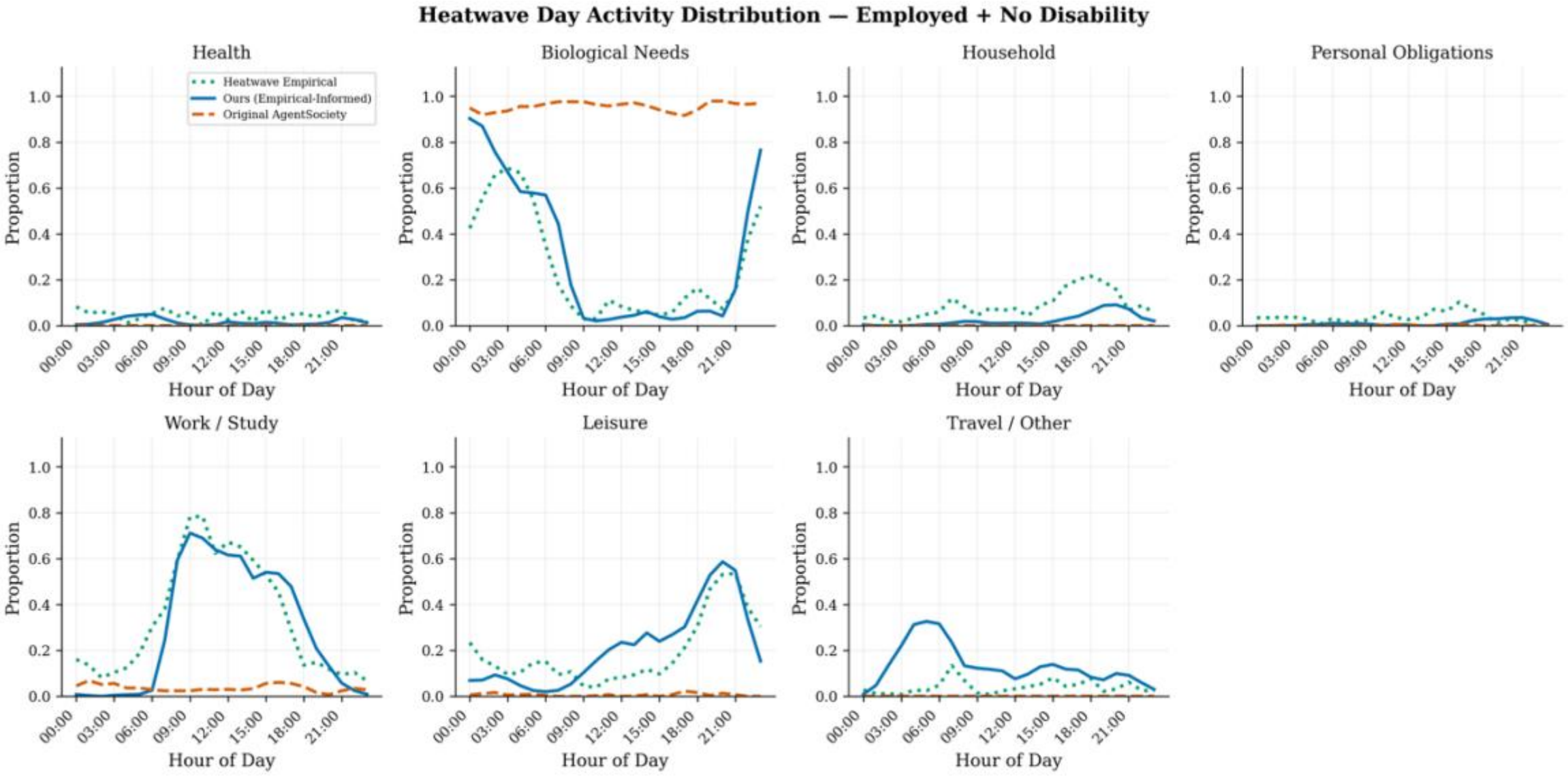

**Fig. S13.** Heatwave-day activity comparison for employed agents without disability.

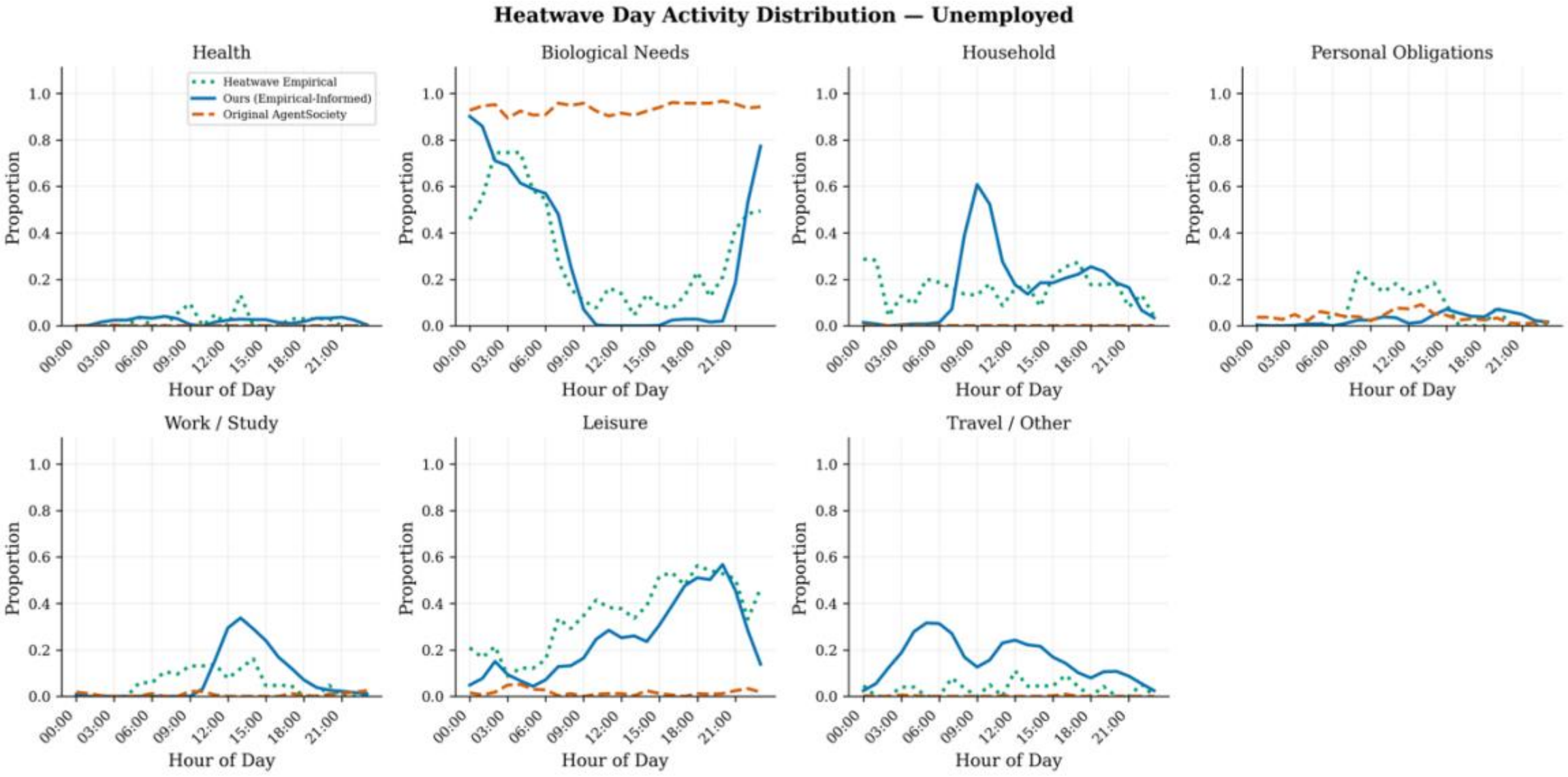

**Fig. S14.** Heatwave activity comparison for unemployed agents.

**Tables**

**Table S1. Normal Day: Simulation vs ATUS Empirical**

| Demographic | Ours (Empirical-Informed) | | | | Original LLM Agent Baseline | | | |
|---|---|---|---|---|---|---|---|---|
| | **MSE** | **MAE** | **Corr** | **JSD** | **MSE** | **MAE** | **Corr** | **JSD** |
| Unemployed | 0.010471 | 0.067 | 0.8587 | 0.010632 | 0.066956 | 0.139 | 0.5476 | 0.036688 |
| Hybrid Emp. + Children | 0.006749 | 0.0513 | 0.9282 | 0.007646 | 0.072449 | 0.1333 | 0.4722 | 0.034997 |
| Hybrid Emp. + No Children | 0.005777 | 0.0471 | 0.9413 | 0.006254 | 0.067697 | 0.1326 | 0.5002 | 0.032456 |
| Hybrid Emp. + Disability | 0.005688 | 0.0484 | 0.9322 | 0.007389 | 0.056946 | 0.1215 | 0.5794 | 0.029694 |
| In-Person Emp. + Children | 0.008118 | 0.0514 | 0.9245 | 0.008 | 0.068716 | 0.1333 | 0.4945 | 0.033636 |
| In-Person Emp. + Disability | 0.011319 | 0.0644 | 0.8798 | 0.010062 | 0.057967 | 0.1285 | 0.5843 | 0.031255 |
| In-Person Emp. + No Children | 0.0101 | 0.0613 | 0.8969 | 0.009808 | 0.059677 | 0.1273 | 0.5988 | 0.029653 |
| Hybrid Emp. + No Disability | 0.005881 | 0.0479 | 0.9397 | 0.006873 | 0.081162 | 0.143 | 0.4324 | 0.038294 |
| In-Person Emp. + No Disability | 0.010996 | 0.0599 | 0.9048 | 0.00943 | 0.062339 | 0.1286 | 0.5452 | 0.031036 |
| **AVERAGE** | **0.008344** | **0.055411** | **0.911789** | **0.008455** | **0.06599** | **0.1319** | **0.528289** | **0.033079** |

**Table S2. Heatwave Day: Simulation vs Heatwave Survey Empirical**

| Demographic | Ours (Empirical-Informed) | | | | | Original LLM Agent Baseline | | | | |
|---|---|---|---|---|---|---|---|---|---|---|
| | MSE | MAE | Corr | JSD | Closer | MSE | MAE | Corr | JSD | Closer |
| Unemployed | 0.016783 | 0.0935 | 0.7451 | 0.017041 | heatwave | 0.08783 | 0.181 | 0.4209 | 0.044119 | normal |
| Employed + Children | 0.012689 | 0.0809 | 0.8418 | 0.012964 | heatwave | 0.112529 | 0.2022 | 0.2267 | 0.051231 | normal |
| Employed + Disability | 0.010033 | 0.0741 | 0.866 | 0.01301 | heatwave | 0.092796 | 0.1877 | 0.3919 | 0.046904 | normal |
| Employed + No Children | 0.011694 | 0.0808 | 0.8544 | 0.012317 | heatwave | 0.086192 | 0.1771 | 0.43 | 0.040098 | normal |
| Employed + No Disability | 0.010476 | 0.0734 | 0.8745 | 0.010971 | heatwave | 0.109989 | 0.1995 | 0.2739 | 0.050177 | normal |
| **AVERAGE** | **0.012335** | **0.080540** | **0.836360** | **0.013261** | **5/5 HW** | **0.097867** | **0.189500** | **0.348680** | **0.046506** | **0/5 HW** |

**Table S3. Heatwave Behavior Change Magnitude**

| Demographic | Ours (Empirical-Informed) Change Mag. | Original LLM Agent Change Mag. | Heatwave Change Mag. | Ours/Emp | Base/Emp |
|---|---|---|---|---|---|
| Unemployed | 0.04455 | 0.01033 | 0.09268 | 0.48 | 0.11 |
| Employed + Children | 0.03565 | 0.01604 | 0.09145 | 0.39 | 0.18 |
| Employed + Disability | 0.03732 | 0.02068 | 0.07701 | 0.48 | 0.27 |
| Employed + No Children | 0.03446 | 0.01674 | 0.07034 | 0.49 | 0.24 |
| Employed + No Disability | 0.03664 | 0.01759 | 0.0763 | 0.48 | 0.23 |
| **AVERAGE** | **0.03772** | **0.01628** | **0.08156** | **0.464** | **0.206** |

**Table S4. Per-Category Behavioral Delta**

| Category | Ours (Empirical-Informed) | | Original LLM Agent Baseline | | Heatwave Change | |
|---|---|---|---|---|---|---|
| | **Mean Δ** | **Mean \|Δ\|** | **Mean Δ** | **Mean \|Δ\|** | **Mean Δ** | **Mean \|Δ\|** |
| Health | 0.00632 | 0.00632 | 0.00033 | 0.00033 | 0.04481 | 0.04481 |
| Biological Needs | -0.06665 | 0.06665 | 0.03793 | 0.03954 | -0.10356 | 0.10356 |
| Household | 0.00483 | 0.00492 | 0 | 0 | 0.04152 | 0.04152 |
| Personal Obligations | 0.0058 | 0.00792 | -0.00801 | 0.00801 | 0.00565 | 0.01559 |
| Work / Study | -0.00402 | 0.01487 | -0.03161 | 0.03307 | 0.02 | 0.0445 |
| Leisure | 0.03702 | 0.03702 | 0.00111 | 0.00166 | 0.04557 | 0.0468 |
| Travel / Other | 0.0167 | 0.02736 | 0.00025 | 0.0003 | -0.01018 | 0.01127 |

**Table S5. Population Segmentation Summary**

| ID | Group label ($G_k$) | Defining ACS attributes | Key routine constraints (weekday) |
|---|---|---|---|
| G1 | Employed, in-person, with children | EMP=employed; WORKMODE=in-person; CHILD<18=Yes | Work block outside home; childcare drop/pick windows; peak shopping trips early evening |
| G2 | Employed, in-person, no children | EMP=employed; in-person; CHILD<18=No | Work block outside home; discretionary/leisure more flexible |
| G3 | Employed, in-person, with disability | EMP=employed; in-person; DISABILITY=Yes | Work block outside home; higher probability of medical/administrative trips; shorter travel durations |
| G4 | Employed, in-person, without disability | EMP=employed; in-person; DISABILITY=No | Standard in-person work profile |
| G5 | Employed, hybrid, with children | EMP=employed; WORKMODE=hybrid | Mix of home/office workdays; childcare windows |
| G6 | Employed, hybrid, no children | EMP=employed; hybrid | Mixture of home/office work; later commute starts on office days |
| G7 | Employed, hybrid, with disability | EMP=employed; hybrid; DISABILITY=Yes | More home-based workdays; fewer/shorter discretionary trips |
| G8 | Employed, hybrid, without disability | EMP=employed; hybrid; DISABILITY=No | Hybrid baseline |
| G9 | Unemployed | EMP=not employed | No work block; higher maintenance/errand probability; greater daytime home presence |

**SI References**